\documentclass{article}

\PassOptionsToPackage{numbers, compress}{natbib}

\usepackage[preprint]{tackling_climate_workshop_style}

\usepackage[utf8]{inputenc} 
\usepackage[T1]{fontenc}    
\usepackage{hyperref}       
\usepackage{url}            
\usepackage{booktabs}       
\usepackage{amsfonts}       
\usepackage{graphicx}       
\usepackage{nicefrac}       
\usepackage{microtype}      
\title{Computer Vision and Knowledge-Grounded Language Models for Intelligent Beekeeper Assistance} 

\author{%
  Asaduz Zaman \\
  Faculty of Information Technology \\
  Monash University \\
  Clayton, VIC-3800, Australia \\
  \texttt{asaduzzaman@monash.edu} \\
  \And
  Mohaimenul Azam Khan Raiaan \\
  Faculty of Information Technology \\
  Monash University \\
  Clayton, VIC-3800, Australia \\
  \texttt{mohaimenul.raiaan@monash.edu} \\
  \And
  Malika Nisal Ratnayake \\
  Faculty of Information Technology \\
  Monash University \\
  Clayton, VIC-3800, Australia \\
  \texttt{malika.ratnayake@monash.edu} \\
}

\begin{document}

\maketitle

\begin{abstract}
Managed and wild bees sustain roughly a third of the global food products, yet colonies face mounting stress from climate-change including extreme weather patterns, altered parasite and disease patterns. Timely assessment of colony condition such as understanding colony strength, brood area, honey and pollen storage, comb usage, is essential to adaptive management. However, manual hive frame inspection is slow, costly, and quality depends on scarce expertise. We present an artificial targeted intelligent system framework focusing on edge deployable pipeline that turns hive frame photographs into interpretable and action-oriented inspection report. The pipeline includes an object detector to detect biologically-meaningful comb-cell classes; detections are distilled into structured fact schema; facts are converted into constrained factual text; the text triggers a grounded knowledge-base for interpretations and recommendations. Knowledge-base entries are required to be triggerable by evidence expressed in the fact schema. We built our system based on open source FAIRHiveFrames-1K dataset and favour reproducible, edge-deployable models over large general purpose frontier models. We additionally release FAIRHiveText-1K dataset that turns FAIRHiveFrames-1K dataset into interchangeable facts and text interpretation of the facts, making it easy for further LLM-based research into this area. We also propose a pathway to develop the grounded knowledge-base that will be helpful in providing interpretable and actionable reports from the facts.
\end{abstract}

\section{Introduction}

Honeybees are essential to global food production and ecosystem services~\citep{potts2016assessment,klein2007importance,shrestha2008honeybees}, and insufficient pollination already limits yields in several major crops ~\citep{reilly2020crop}. Climate change further exposes colonies to extreme and unstable weather, shifting seasons, altered floral resources, and changing pest, parasite, and diseases, with weather instability linked to colony losses through interactions with parasite pressure~\citep{vasiliev2021role,bonebrake2012climate,wilfert2016deformed, insolia2022honey}. As these conditions become less predictable, management based on seasonal calendars becomes less reliable, increasing the need for precision apiculture guided by timely, colony-specific assessments.

Hive-frame inspection provides important indicators of colony condition, including brood development, food reserves, colony strength, and comb use~\citep{alves2020automatic}. While precision apiculture can continuously monitor colony-level signals such as weight, temperature, and acoustics,~\citep{zacepins2015challenges}, frame-level indicators still require expert visual assessment, limiting the scale and frequency of assessment. Computer vision can automate their extraction~\citep{kulyukin2026fairhiveframes}, but these predictions alone do not tell beekeepers what observations imply or what actions to take. This requires connecting visual evidence with domain knowledge in a traceable manner.

Large language models (LLMs) offer a natural interface for translating such observations into personalised advice~\citep{tzachor2023large}, but hallucination limits their reliability for management recommendations~\citep{ji2023survey}. Retrieval-augmented generation (RAG) mitigates this by grounding outputs in an explicit, inspectable knowledge-base ~\citep{lewis2020retrieval}.

We therefore propose an interpretable, edge-deployable AI framework that separates visual perception, factual representation, knowledge retrieval, and language generation. Computer-vision predictions are transformed into structured facts that support retrieval of relevant domain knowledge and evidence-grounded generation of interpretations and recommendations. We instantiate the framework for honeycomb-cell analysis using FAIRHiveFrames-1K~\cite{kulyukin2026fairhiveframes}, an open-source dataset of 1,265 expert-annotated hive-frame images, and introduce FAIRHiveText-1K, which extends these visual annotations into structured facts and textual descriptions for grounded language-based analysis. Together, they provide a concrete baseline of the proposed framework towards scalable and accessible decision support for adaptive honeybee management under increasing climate variability.


\section{Proposed Approach}
\begin{figure}
    \centering
    \includegraphics[width=0.7\linewidth]{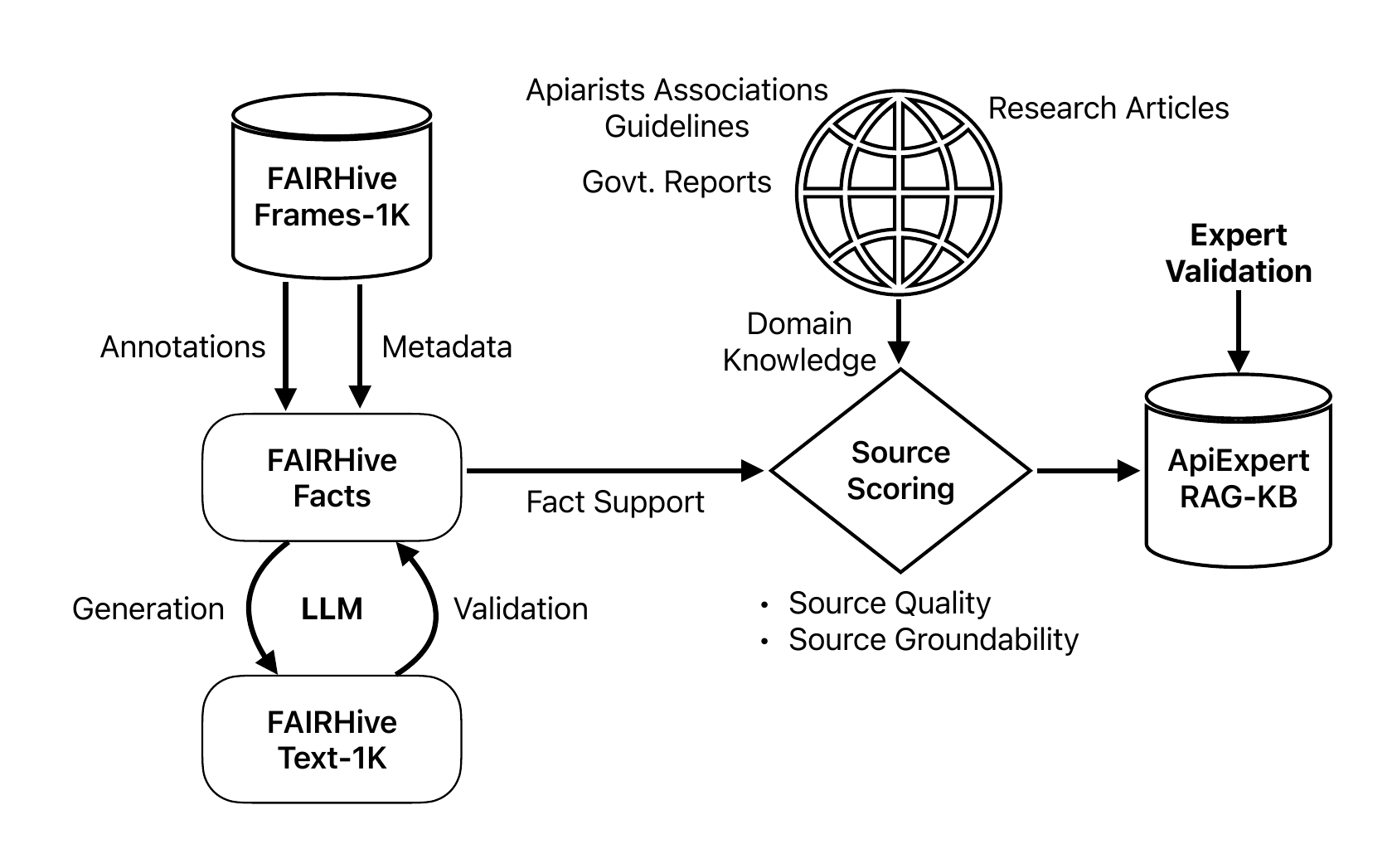}
    \caption{Knowledge-base construction. Annotations from FAIRHiveFrames-1K yield structured facts, which an LLM expands into FAIRHiveText-1K with round-trip validation; external domain knowledge is scored for quality and fact-schema groundability, then expert-validated into the ApiExpert RAG-KB.}
    \label{fig:framework}
\end{figure}

\begin{figure}
    \centering
    \includegraphics[width=0.95\linewidth]{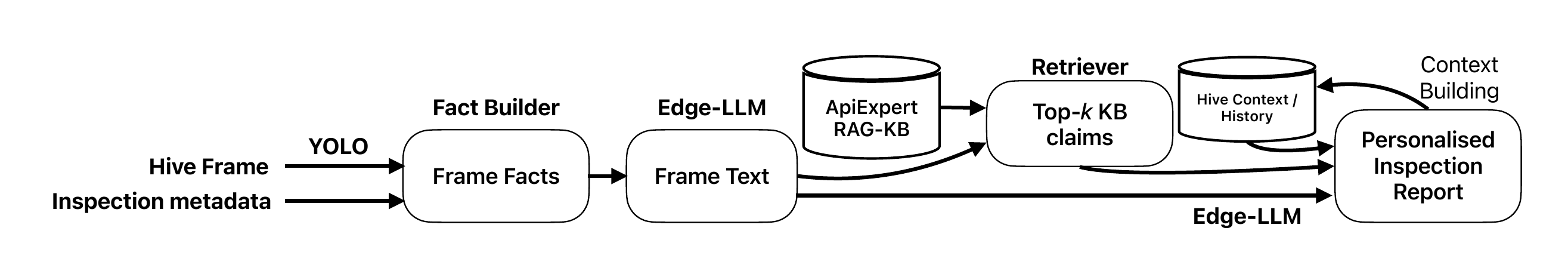}
    \caption{On-device inference pipeline. A YOLO-based detector processes a hive-frame image, and a fact builder combines its prediction with inspection metadata into structured frame facts, which an Edge-LLM renders as a frame description. The description retrieves the top-$k$ supporting claims from the ApiExpert RAG-KB (Fig.~\ref{fig:framework}), and the same model generates a personalised inspection report conditioned on the frame description, retrieved claims, and hive context and history. Completed reports are stored in the hive history, informing subsequent inspection.}
    \label{fig:pipeline}
\end{figure}

We propose a two-stage framework (Figs.~\ref{fig:framework} and~\ref{fig:pipeline}) that transforms computer-vision observations from honeybee comb-frame into personalised, knowledge-grounded information for beekeepers, extending prior work on automated and sensor-based hive monitoring~\cite{alves2020automatic,zacepins2015challenges,zaman2023framework}. The framework separates knowledge-base construction from operational inference. In the first stage, structured and textual representations of hive-frame information are developed alongside a curated apiculture knowledge base. In the second stage, these resources are integrated with computer-vision predictions, inspection metadata, and hive-specific context to generate personalised inspection reports.


\subsection{Stage 1: Knowledge-Base Construction}
\label{stage1}
The first stage (Fig.~\ref{fig:framework}) converts the FAIRHiveFrames-1K annotations and associated metadata into structured and textual factual representations. Each FAIRHiveFrames-1K sample comprises a comb-frame image, associated annotations, metadata, and a unique identifier. The existing cell-level annotations classify honeycomb cells into six content categories~\cite{alves2020automatic,kulyukin2026fairhiveframes}. These cell-level annotations are aggregated deterministically into frame-level measurements, such as cell counts and class proportions, forming \textit{FAIRHiveFacts}. A pretrained LLM with a constrained prompt template renders these facts as \textit{FAIRHiveText-1K}: concise factual descriptions of comb composition, deliberately excluding diagnoses or recommendations. Generation validation is performed by extracting facts from the generated text and comparing them with the original structured facts, providing a mechanism to assess consistency between the two representations. In parallel, external apiculture knowledge is collected from sources including apiarist association guidelines, government reports, and relevant research articles. Candidate information is assessed according to \emph{source quality} and \emph{source groundability}, where groundability reflects whether the knowledge can be connected to observations represented within the FAIRHiveFacts schema. Following expert validation, the selected domain knowledge is incorporated into the \textit{ApiExpert RAG-KB}. This process produces a curated knowledge resource designed to support the retrieval of information relevant to observations obtained from hive-frame analysis.
\subsection{Stage 2: Personalised Beekeeper Assistance}
\label{stage2}
The second stage (Fig.~\ref{fig:pipeline}) applies the constructed resources during hive inspection. A YOLO-based detector processes an input hive-frame image, and a Fact Builder combines the resulting predictions with inspection metadata to construct structured \textit{Frame Facts}. An Edge-LLM then converts these facts into a concise \textit{Frame Text} representation. Subsequently, a retriever uses the Frame Text to obtain the top-$k$ relevant claims from the \textit{ApiExpert RAG-KB} developed in Stage~1 (Sect. \ref{stage1}). These retrieved claims are supplied to the Edge-LLM together with the Frame Text and available hive context and history to generate a personalised inspection report~\cite{lewis2020retrieval}. In this way, the generation of reports is conditioned on both observations derived from the current hive frame and relevant external apiculture knowledge. The history and context of the hive are maintained through a context-building component. Completed personalised inspection reports are added to the hive history, allowing information accumulated from previous inspections to contribute to the context available during subsequent analyses. The resulting pipeline therefore combines computer-vision observations, retrieved domain knowledge, inspection metadata, and accumulated hive information to support personalised, knowledge-grounded beekeeper assistance.

\section{Pathway to Impact and Future Work}

The proposed framework provides a pathway from automated hive observations to accessible, evidence-grounded decision support, enabling beekeepers to identify changes in colony condition and respond more effectively to increasing climate variability. Future work will implement and evaluate the complete pipeline, including edge deployment, knowledge-base construction, report generation, and integration of hive and environmental context. Ultimately, linking repeated colony assessments with climatic conditions could support earlier identification of climate-related stress and more adaptive, data-driven beekeeping practices.

{
\small
\bibliographystyle{unsrtnat}
\bibliography{references}
}






\end{document}